\documentclass[11pt,a4paper]{article}
\usepackage[hyperref]{acl2019}
\usepackage{times}
\usepackage{latexsym}
\usepackage{todonotes}
\usepackage{ulem}
\usepackage{multirow}
\usepackage{amssymb}
\usepackage{paralist}
\usepackage{amsmath}
\usepackage{soul}
\usepackage{booktabs}
\DeclareMathOperator*{\argmax}{arg\,max}

\usepackage{url}

\aclfinalcopy 
\def\aclpaperid{634} 

\title{Unsupervised Question Answering by Cloze Translation}

\author{Patrick Lewis \\
  Facebook AI Research \\
  University College London \\
  \texttt{plewis@fb.com} \\\And
  Ludovic Denoyer \\
  Facebook AI Research  \\
  \texttt{denoyer@fb.com} \\\And
  Sebastian Riedel \\
  Facebook AI Research  \\
  University College London \\
  \texttt{sriedel@fb.com} \\}

\date{}

\begin{document}
\normalem

\maketitle
\begin{abstract}

Obtaining training data for Question Answering (QA) is time-consuming and resource-intensive, and existing QA datasets are only available for limited domains and languages. In this work, we explore to what extent high quality training data is \emph{actually} required for Extractive QA, and investigate the possibility of \emph{unsupervised} Extractive QA. We approach this problem by first learning to generate context, question and answer triples in an unsupervised manner, which we then use to synthesize Extractive QA training data automatically. To generate such triples, we first sample random context paragraphs from a large corpus of documents and then random noun phrases or named entity mentions from these paragraphs as answers.
Next we convert answers in context to ``fill-in-the-blank" cloze questions and finally translate them into natural questions. We propose and compare various unsupervised ways to perform cloze-to-natural question translation, including training an unsupervised NMT model using \emph{non-aligned} corpora of natural questions and cloze questions as well as a rule-based approach. We find that modern QA models can learn to answer human questions surprisingly well using only synthetic training data. We demonstrate that, without using the SQuAD training data \emph{at all}, our approach achieves 
56.4 F1 on SQuAD v1 (64.5 F1 when the answer is a Named entity mention), outperforming early supervised models.

\end{abstract}

\section{Introduction}

Extractive Question Answering (EQA) is the task of answering questions given a context document under the assumption that answers are spans of tokens within the given document. There has been substantial progress in this task in English. For SQuAD~\cite{rajpurkar_squad:_2016}, a common EQA benchmark dataset, current models beat human performance; For SQuAD 2.0~\cite{rajpurkar_know_2018}, ensembles based on BERT~\cite{devlin_bert:_2018} now match human performance. Even for the recently introduced Natural Questions corpus~\cite{kwiatkowski_natural_2019}, human performance is already in reach. In all these cases, very large amounts of training data are available. But, for new domains (or languages), collecting such training data is not trivial and can require significant resources. What if no training data was available \emph{at all}?

\begin{figure}
\centering
\includegraphics[width=0.475\textwidth]{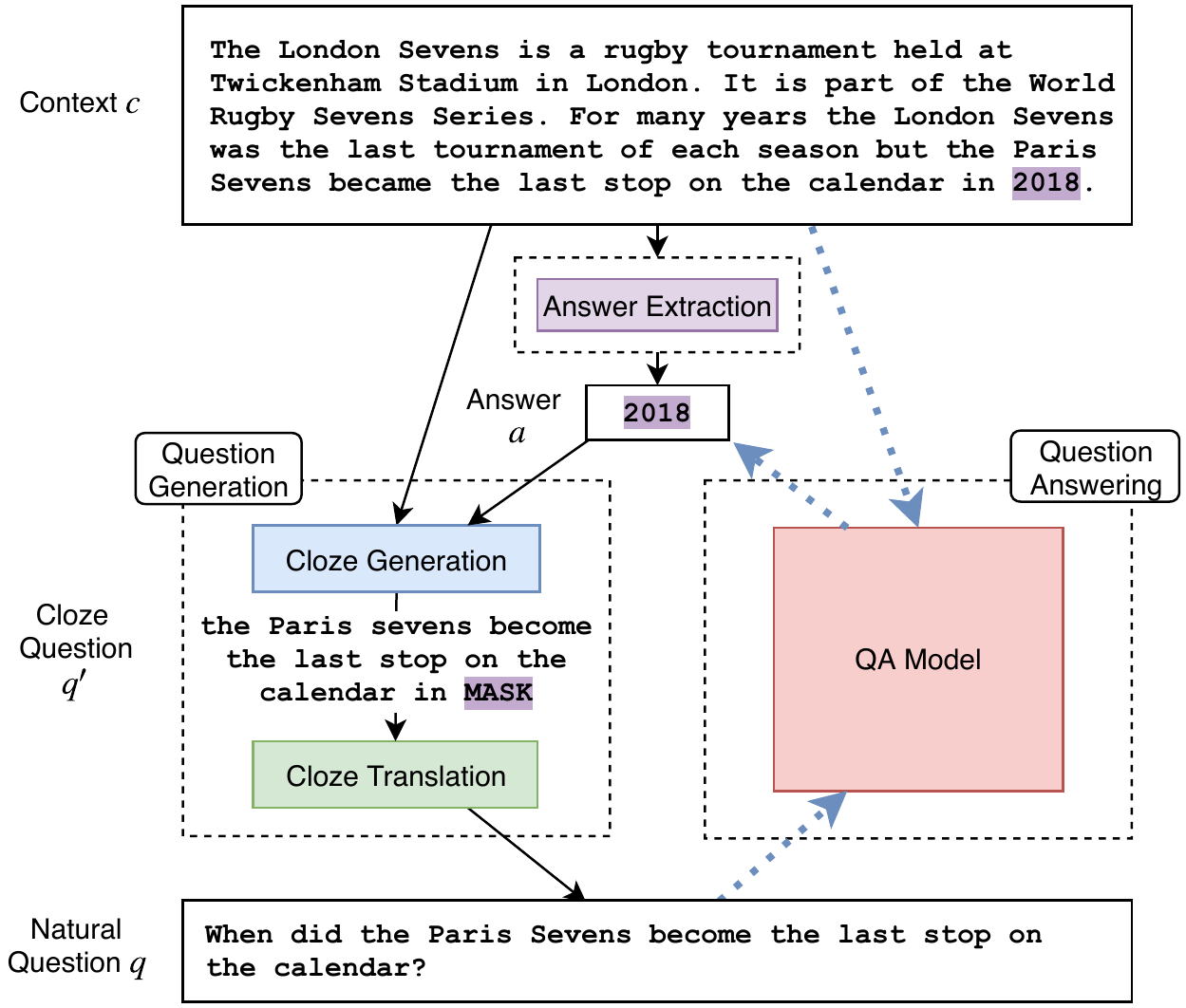}
\caption{A schematic of our approach. The right side (dotted arrows) represents traditional EQA. We introduce unsupervised data generation (left side, solid arrows), which we use to train standard EQA models}
\label{main_diagram}
\end{figure}

In this work we address the above question by exploring the idea of \emph{unsupervised} EQA, a setting in which no aligned question, context and answer data is available. We propose to tackle this by reduction to unsupervised question generation: If we had a method, without using QA supervision, to generate accurate questions given a context document, we could train a QA system using the generated questions. This approach allows us to directly leverage progress in QA, such as model architectures and pretraining routines. This framework is attractive in both its flexibility and extensibility. In addition, our method can also be used to generate \emph{additional} training data in semi-supervised settings.       

Our proposed method, shown schematically in Figure~\ref{main_diagram}, generates EQA training data in three steps. 1) We first sample a paragraph in a target domain---in our case, English Wikipedia. 2) We sample from a set of candidate answers within that context, using pretrained components (NER or noun chunkers) to identify such candidates. 
These require supervision, but no aligned (question, answer) or (question, context) data.
Given a candidate answer and context, we can extract ``fill-the-blank" cloze questions 3) Finally, we convert cloze questions into natural questions using an unsupervised cloze-to-natural question translator.

The conversion of cloze questions into natural questions is the most challenging of these steps. While there exist sophisticated rule-based systems~\cite{heilman_good_2010} to transform statements into questions (for English), we find their performance to be empirically weak for QA~(see Section \ref{experimental}). Moreover, for specific domains or other languages, a substantial engineering effort will be required to develop similar algorithms. Also, whilst supervised models exist for this task, they require the type of annotation unavailable in this setting~(\citealt{du_learning_2017,du_harvesting_2018,hosking_evaluating_2019}, \emph{inter alia}). We overcome this issue by leveraging recent progress in \emph{unsupervised machine translation}~\cite{lample_phrase-based_2018, lample_unsupervised_2017,lample_cross-lingual_2019,ArtetxeLA18}. In particular, we collect a large corpus of natural questions and an \emph{unaligned} corpus of cloze questions, and train a seq2seq model to map between natural and cloze question domains using a combination of online back-translation and de-noising auto-encoding.

In our experiments, we find that in conjunction with the use of modern QA model architectures, unsupervised QA can lead to performances surpassing early supervised approaches~\cite{rajpurkar_squad:_2016}.
We show that forms of cloze ``translation'' that produce (unnatural) questions via word removal and flips of the cloze question lead to better performance than an informed rule-based translator. Moreover, the unsupervised seq2seq model outperforms both the noise and rule-based system. We also demonstrate that our method can be used in a few-shot learning setting, for example obtaining 59.3 F1 with 32 labelled examples, compared to 40.0 F1 without our method.

To summarize, this paper makes the following contributions:
\begin{inparaenum}[i)]
\item The first approach for unsupervised QA, reducing the problem to unsupervised cloze translation, using methods from unsupervised machine translation 
\item Extensive experiments testing the impact of various cloze question translation algorithms and assumptions
\item Experiments demonstrating the application of our method for few-shot learning in EQA.\footnote{Synthetic EQA training data and models that generate it will be made publicly available at \url{https://github.com/facebookresearch/UnsupervisedQA}}
\end{inparaenum}

\section{Unsupervised Extractive QA}\label{UQA}
\newcommand{\context}{\ensuremath{c}}
\newcommand{\question}{\ensuremath{q}}
\newcommand{\answer}{\ensuremath{a}}
\newcommand{\cloze}{\ensuremath{q'}}
\newcommand{\qtoc}{\ensuremath{g_{\question \rightarrow \cloze}}}

We consider \emph{extractive} QA where we are given a question \question{} and a context paragraph \context{} and need to provide an answer $\answer=(b,e)$ with beginning $b$ and end $e$ character indices in \context. Figure~\ref{main_diagram} (right-hand side) shows a schematic representation of this task.

We propose to address unsupervised QA in a two stage approach. We first develop a generative model $p(\question,\answer,\context)$ using no (QA) supervision, and then train a discriminative model $p_r(\answer|\question,\context)$ using $p$ as training data generator. The generator $p(\question,\answer,\context)=p(\context) p(\answer|\context) p(\question|\answer,\context)$ will generate data in a ``reverse direction", first sampling a context via $p(\context)$, then an answer within the context via $p(\answer|\context)$ and finally a question for the answer and context via $p(\question|\answer,\context)$. In the following we present variants of these components. 

\subsection{Context and Answer Generation}

Given a corpus of documents our context generator $p(\context)$ uniformly samples a paragraph \context{} of appropriate length from any document, and the answer generation step creates answer spans \answer{} for \context{} via $p(\answer|\context)$. This step incorporates prior beliefs about what constitutes good answers. We propose two simple variants for $p(\answer|\context)$:

\paragraph{Noun Phrases} We extract all noun phrases from paragraph \context{} and sample uniformly from this set to generate a possible answer span. This requires a chunking algorithm for our language and domain.

\paragraph{Named Entities} We can further restrict the possible answer candidates and focus entirely on named entities. Here we extract all named entity mentions using an NER system and then sample uniformly from these. Whilst this reduces the variety of questions that can be answered, it proves to be empirically effective as discussed in Section \ref{ablations}.  

\subsection{Question Generation}
Arguably, the core challenge in QA is modelling the relation between question and answer. This is captured in the question generator $p(\question|\answer,\context)$ that produces questions from a given answer in context.  We divide this step into two steps: cloze generation $\cloze=\text{cloze}(a,c)$ and translation, $p(\question|\cloze)$. 

\subsubsection{Cloze Generation}
Cloze questions are statements with the answer masked.
In the first step of cloze generation, we reduce the scope of the context to roughly match the level of detail of actual questions in extractive QA. A natural option is the sentence around the answer. Using the context and answer from Figure~\ref{main_diagram}, this might leave us with the sentence \emph{``For many years the London Sevens was the last tournament of each season but the Paris Sevens became the last stop on the calendar in \underline{\hspace{2em}}''}. We can further reduce length by restricting to sub-clauses around the answer, based on access to an English syntactic parser, leaving us with \emph{``the Paris Sevens became the last stop on the calendar in \underline{\hspace{2em}}"}.

\subsubsection{Cloze Translation}
\label{cloze_trans}
Once we have generated a cloze question $\cloze$ we translate it into a form closer to what we expect in real QA tasks. We explore four approaches here.

\paragraph{Identity Mapping}
We consider that cloze questions themselves provide a signal to learn some form of QA behaviour. To test this hypothesis, we use the identity mapping as a baseline for cloze translation. To produce ``questions'' that use the same vocabulary as real QA tasks, we replace the mask token with a wh* word (randomly chosen or with a simple heuristic described in Section \ref{UAQ}).

\paragraph{Noisy Clozes}
One way to characterize the difference between cloze and natural questions is as a form of perturbation. To improve robustness to pertubations, we can inject noise into cloze questions. We implement this as follows. First we delete the mask token from cloze $\cloze$, apply a simple noise function from \citet{lample_phrase-based_2018}, and prepend a wh* word (randomly or with the heuristic in Section \ref{UAQ}) and append a question mark. The noise function consists of word dropout, word order permutation and word masking. The motivation is that, at least for SQuAD, it may be sufficient to simply learn a function to identify a span surrounded by high n-gram overlap to the question, with a tolerance to word order perturbations.

\paragraph{Rule-Based}
Turning an answer embedded in a sentence into a $(\question{},\answer{})$ pair can be understood as a syntactic transformation with wh-movement and a type-dependent choice of wh-word. For English, off-the-shelf software exists for this purpose. We use the popular statement-to-question generator from \citet{heilman_good_2010} which uses a set of rules to generate many candidate questions, and a ranking system to select the best ones.

\paragraph{Seq2Seq}
The above approaches either require substantial engineering and prior knowledge (rule-based) or are still far from generating natural-looking questions (identity, noisy clozes). We propose to overcome both issues through unsupervised training of a seq2seq model that translates between cloze and natural questions. More details of this approach are in Section~\ref{UAQ}. 

\subsection{Question Answering}
\label{qa}
Extractive Question Answering amounts to finding the best answer \answer{} given question \question{} and context \context{}.  We have at least two ways to achieve this using our generative model:

\paragraph{Training a separate  QA system} 
The generator is a source of training data for any QA architecture at our disposal. Whilst the data we generate is unlikely to match the quality of real QA data, we hope QA models will learn basic QA behaviours.

\paragraph{Using Posterior}\label{using_posterior}
Another way to extract the answer is to find \answer{} with the highest posterior $p(\answer|\context,\question)$. Assuming uniform answer probabilities conditioned on context $p(\answer|\context)$, this amounts to calculating $\argmax_{\answer'} p(\question|\answer',\context)$ by testing how likely each possible candidate answer could have generated the question, a similar method to the supervised approach of \citet{lewis_generative_2018_2}.

\subsection{Unsupervised Cloze Translation}\label{UAQ}
To train a seq2seq model for cloze translation we borrow ideas from recent work in unsupervised Neural Machine Translation~(NMT). At the heart of most these approaches are \emph{nonparallel} corpora of source and target language sentences. In such corpora, no source sentence has any translation in the target corpus and vice versa. Concretely, in our setting, we aim to learn a function which maps between the question (target) and cloze question (source) domains without requiring aligned corpora. For this, we need large corpora of cloze questions $C$ and natural questions $Q$.

\paragraph{Cloze Corpus} We create the cloze corpus $C$ by applying the procedure outlined in Section \ref{cloze_trans}. Specifically we consider Noun Phrase (NP) and Named Entity mention (NE) answer spans, and cloze question boundaries set either by the sentence or sub-clause that contains the answer.\footnote{We use SpaCy for Noun Chunking and NER, and AllenNLP for the \citet{stern_minimal_2017} parser.}
We extract 5M cloze questions from randomly sampled wikipedia paragraphs, and build a corpus $C$ for each choice of answer span and cloze boundary technique. Where there is answer entity typing information (i.e. NE labels), we use type-specific mask tokens to represent one of 5 high level answer types. See Appendix \ref{cloze_feat} for further details. 

\paragraph{Question Corpus} We mine questions from English pages from a recent dump of common crawl using simple selection criteria:\footnote{\url{http://commoncrawl.org/}}  We select sentences that start in one of a few common wh* words, (``\texttt{how much}", ``\texttt{how many}", ``\texttt{what}", ``\texttt{when}", ``\texttt{where}" and ``\texttt{who}") and end in a question mark. We reject questions that have repeated question marks or ``?!", or are longer than 20 tokens. This process yields over 100M english questions when deduplicated. Corpus $Q$ is created by sampling 5M questions such that there are equal numbers of questions starting in each wh* word.  

Following \citet{lample_phrase-based_2018}, we use $C$ and $Q$ to train translation models $p_{s \rightarrow t}(\question | \cloze)$ and $p_{t \rightarrow s}(\cloze|\question)$ which translate cloze questions into natural questions and vice-versa. This is achieved by a combination of in-domain training via denoising autoencoding and cross-domain training via online-backtranslation. This could also be viewed as a style transfer task, similar to \citet{subramanian_multiple-attribute_2018}. At inference time, `natural' questions are generated from cloze questions as $\argmax_\question p_{s \rightarrow t} (\question | \cloze)$.\footnote{We also experimented with language model pretraining in a method similar to \citet{lample_cross-lingual_2019}. Whilst generated questions were generally more fluent and well-formed,  we did not observe significant changes in QA performance. Further details in Appendix \ref{xlm}} Further experimental detail can be found in Appendix \ref{uqa_training_details}.

\paragraph{Wh* heuristic} 

In order to provide an appropriate wh* word for our ``identity" and ``noisy cloze" baseline question generators, we introduce a simple heuristic rule that maps each answer type to the most appropriate  wh* word. For example, the ``\texttt{TEMPORAL}" answer type is mapped to ``\texttt{when}". During experiments, we find that the unsupervised NMT translation functions sometimes generate inappropriate wh* words for the answer entity type, so we also experiment with applying the wh* heuristic to these question generators. For the NMT models, we apply the heuristic by prepending target questions with the answer type token mapped to their wh* words at training time. E.g. questions that start with ``\texttt{when}" are prepended with the token ``\texttt{TEMPORAL}". Further details on the wh* heuristic are in Appendix \ref{uqa_training_details_wh_heuristic}.  

\section{Experiments}
\label{experimental}

We want to explore what QA performance can be achieved without using \emph{aligned} \question, \answer{} data, and how this compares to supervised learning and other approaches which do not require training data. Furthermore, we seek to understand the impact of different design decisions upon QA performance of our system and to explore whether the approach is amenable to few-shot learning when only a few \question{},\answer{} pairs are available. Finally, we also wish to assess whether unsupervised NMT can be used as an effective method for question generation.

\subsection{Unsupervised QA Experiments}

For the synthetic dataset training method, we consider two QA models: finetuning BERT \cite{devlin_bert:_2018} and BiDAF + Self Attention \cite{clark_simple_2017}.\footnote{We use the HuggingFace implementation of BERT, available at \url{https://github.com/huggingface/pytorch-pretrained-BERT}, and the documentQA implementation of BiDAF+SA, available at \url{https://github.com/allenai/document-qa}} For the posterior maximisation method, we extract cloze questions from both sentences and sub-clauses, and use the NMT models to estimate $p(\question|\context,\answer)$.  We evaluate using the standard Exact Match (EM) and F1 metrics.

As we cannot assume access to a development dataset when training unsupervised models, the QA model training is halted when QA performance on a held-out set of synthetic QA data plateaus. We do, however, use the SQuAD development set to assess which model components are important (Section \ref{ablations}). 
To preserve the integrity of the SQuAD test set, we only submit our best performing system to the test server. 

We shall compare our results to some published baselines. \citet{rajpurkar_squad:_2016} use a supervised logistic regression model with feature engineering, and a sliding window approach that finds answers using word overlap with the question. \citet{kaushik_how_2018} train (supervised) models that disregard the input question and simply extract the most likely answer span from the context. To our knowledge, ours is the first work to deliberately target unsupervised QA on SQuAD. \citet{dhingra_simple_2018} focus on semi-supervised QA, but do publish an unsupervised evaluation. To enable fair comparison, we re-implement their approach using their publicly available data, and train a variant with BERT-Large.\footnote{\url{http://bit.ly/semi-supervised-qa}} Their approach also uses cloze questions, but without translation, and heavily relies on the structure of wikipedia articles.

\begin{table}
\centering
\small
\begin{tabular}{ p{5.2cm} p{0.5cm}  p{0.5cm} }
\textbf{Unsupervised Models} &EM&F1 \\
\toprule
BERT-Large Unsup. QA (ens.) & \textbf{47.3} & \textbf{56.4} \\
BERT-Large Unsup. QA (single) & 44.2 & 54.7 \\
BiDAF+SA \cite{dhingra_simple_2018} & 3.2$^\dagger$ & 6.8$^\dagger$\\
BiDAF+SA \cite{dhingra_simple_2018}$^\ddagger$ & 10.0* & 15.0*\\
BERT-Large \cite{dhingra_simple_2018}$^\ddagger$  & 28.4* & 35.8*\\
\midrule
\textbf{Baselines} & EM & F1 \\\toprule 
Sliding window \cite{rajpurkar_squad:_2016} & \textbf{13.0} & \textbf{20.0} \\
Context-only \cite{kaushik_how_2018}  & 10.9 & 14.8 \\
Random \cite{rajpurkar_squad:_2016}   & 1.3  & 4.3 \\
\midrule
\textbf{Fully Supervised Models} & EM & F1 \\\toprule
BERT-Large \cite{devlin_bert:_2018} & \textbf{84.1} & \textbf{90.9} \\
BiDAF+SA \cite{clark_simple_2017} & 72.1 & 81.1 \\
Log. Reg. + FE \cite{rajpurkar_squad:_2016}& 40.4 & 51.0 \\

\end{tabular}
\caption{Our best performing unsupervised QA models compared to various baselines and supervised models. * indicates results on SQuAD dev set. $\dagger$ indicates results on non-standard test set created by \citet{dhingra_simple_2018}. $\ddagger$ indicates our re-implementation}
\label{main_table}
\end{table}

Our best approach attains 54.7 F1 on the SQuAD test set; an ensemble of 5 models (different seeds) achieves 56.4 F1. Table~\ref{main_table} shows the result in context of published baselines and supervised results. Our approach significantly outperforms baseline systems and \citet{dhingra_simple_2018} and surpasses early supervised methods.

\subsection{Ablation Studies and Analysis}
\label{ablations}

\begin{table*}
\small
\centering
\begin{tabular}{ c  c  c  c  c  c | c  c | c  c }

\multirow{2}{30pt}{\centering \textbf{Cloze Answer}} & \multirow{2}{40pt}{\centering \textbf{Cloze Boundary}} & \multirow{2}{45pt}{\centering \textbf{Cloze Translation}} & \multirow{2}{37pt}{\centering \textbf{Wh* Heuristic}} &  \multicolumn{2}{c}{\textbf{BERT-Base}} & \multicolumn{2}{c}{\textbf{BiDAF+SA}} & \multicolumn{2}{c}{\textbf{Posterior Max.}}\\
  &&&&EM&\multicolumn{1}{c}{F1}&\multicolumn{1}{c}{EM}&\multicolumn{1}{c}{F1}&\multicolumn{1}{c}{EM}&F1\\
  \toprule
  NE & Sub-clause & UNMT & \checkmark  & \textbf{38.6} & \textbf{47.8} & \textbf{32.3} & \textbf{41.2} & \textbf{17.1} & \textbf{21.7}\\
  NE & Sub-clause & UNMT & $\times$    & 36.9 & 46.3 & 30.3 & 38.9 & 15.3 & 19.8 \\
  NE & Sentence   & UNMT & $\times$    & 32.4 & 41.5 & 24.7 & 32.9 & 14.8 & 19.0 \\
  NP & Sentence   & UNMT & $\times$    & 19.8 & 28.4 & 18.0 & 26.0 & 12.9 & 19.2\\
  \midrule
  NE & Sub-clause & Noisy Cloze & \checkmark & 36.5 & 46.1 & 29.3 & 38.7 & - & - \\
  NE & Sub-clause & Noisy Cloze & $\times$   & 32.9 & 42.1 & 26.8 & 35.4 & - & -\\
  NE & Sentence   & Noisy Cloze & $\times$   & 30.3 & 39.5 & 24.3 & 32.7 & - & - \\
  NP & Sentence   & Noisy Cloze & $\times$   & 19.5 & 29.3 & 16.6 & 25.7 & - & -\\
  \midrule
  NE & Sub-clause & Identity & \checkmark &  24.2 & 34.6 & 12.6 & 21.5 & - & -\\
  NE & Sub-clause & Identity & $\times$   &  21.9 & 31.9 & 16.1 & 26.8 & - & -\\
  NE & Sentence   & Identity & $\times$   &  18.1 & 27.4 & 12.4 & 21.2 & - & -\\
  NP & Sentence   & Identity & $\times$   &  14.6 & 23.9 & 6.6 & 13.5 & - & -\\
  \midrule
   \multicolumn{4}{c}{Rule-Based \cite{heilman_good_2010}} &  16.0 & 37.9 & 13.8 & 35.4 & - & - \\
\bottomrule
\end{tabular}
\caption{Ablations on the SQuAD development set. ``Wh* Heuristic" indicates if a  heuristic was used to choose sensible Wh* words during cloze translation. NE and NP refer to named entity mention and noun phrase answer generation.}

\label{ablations_table}
\end{table*}

To understand the different contributions to the performance, we undertake an ablation study. All ablations are evaluated using the SQUAD development set. We ablate using BERT-Base and BiDAF+SA, and our best performing setup is then used to fine-tune a final BERT-Large model, which is the model in Table~\ref{main_table}. All experiments with BERT-Base were repeated with 3 seeds to account for some instability encountered in training; we report mean results. Results are shown in Table~\ref{ablations_table}, and observations and aggregated trends are highlighted below.

\paragraph{Posterior Maximisation vs. Training on generated data} Comparing Posterior Maximisation with BERT-Base and BiDAF+SA columns in Table~\ref{ablations_table} shows that training QA models is more effective than maximising question likelihood. As shown later, this could partly be attributed to QA models being able to generalise answer spans, returning answers at test-time that are not always named entity mentions.
BERT models also have the advantage of linguistic pretraining, further adding to generalisation ability.

\paragraph{Effect of Answer Prior} Named Entities (NEs) are a more effective answer prior than noun phrases (NPs). Equivalent BERT-Base models trained with NEs improve on average by 8.9 F1 over NPs.  \citet{rajpurkar_squad:_2016} estimate 52.4\% of answers in SQuAD are NEs, whereas (assuming NEs are a subset of NPs), 84.2\% are NPs. However, we found that there are on average 14 NEs per context compared to 33 NPs, so using NEs in training may help reduce the search space of possible answer candidates a model must consider.

\begin{figure}
\centering
  \includegraphics[width=0.48\textwidth]{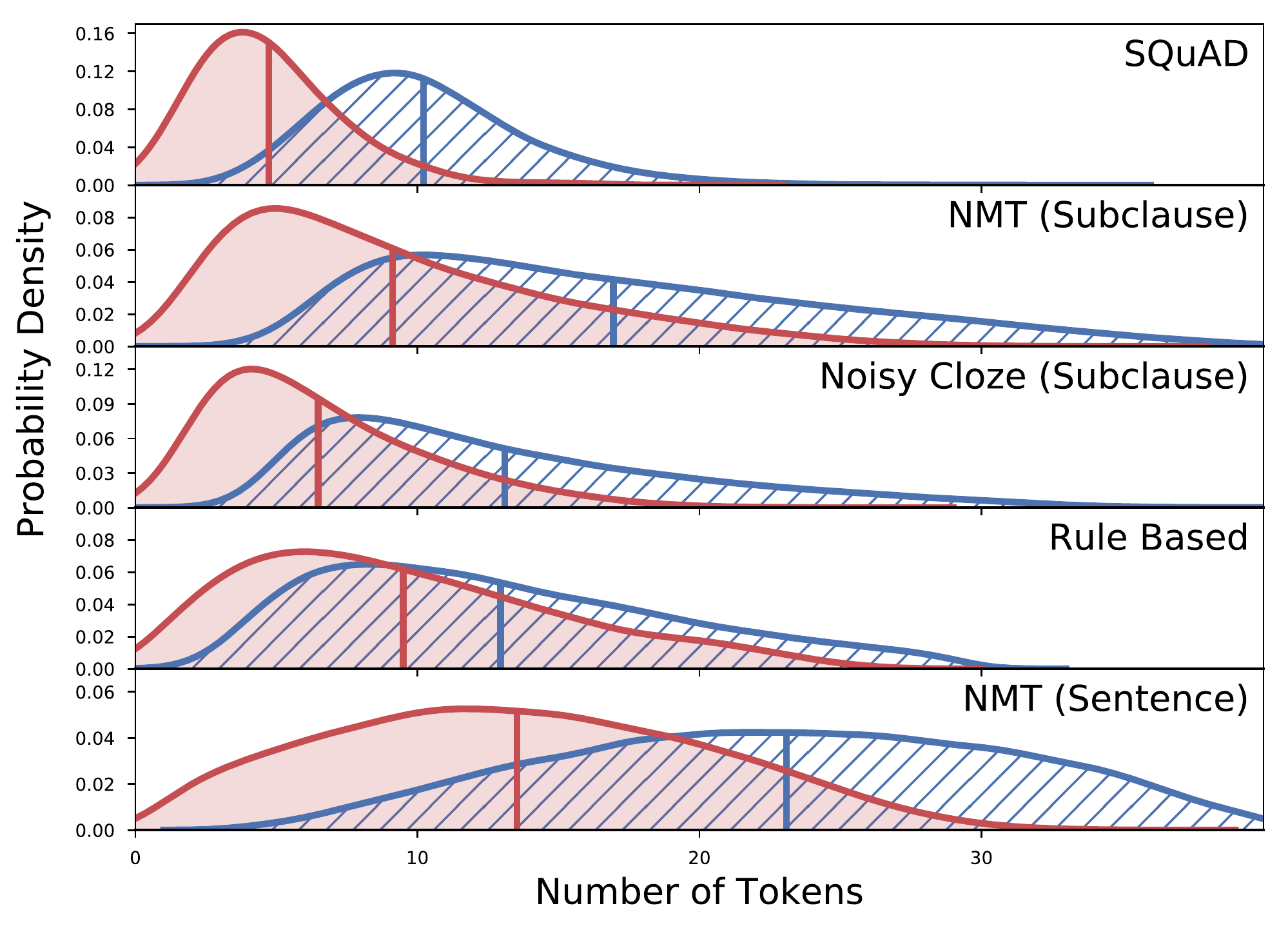}
  \caption{Lengths (blue, hashed) and longest common subsequence with context (red, solid) for SQuAD questions and various question generation methods.}
  \label{question_lengths}
\end{figure}

\paragraph{Effect of Question Length and Overlap} As shown in Figure~\ref{question_lengths}, using sub-clauses for generation leads to shorter questions and shorter common subsequences to the context, which more closely match the distribution of SQuAD questions. Reducing the length of cloze questions helps the translation components produce simpler, more precise questions. Using sub-clauses leads to, on average +4.0 F1 across equivalent sentence-level BERT-Base models. The ``noisy cloze" generator produces shorter questions than the NMT model due to word dropout, and shorter common subsequences due to the word  perturbation noise.

\paragraph{Effect of Cloze Translation} Noise acts as helpful regularization when comparing the ``identity" cloze translation functions to ``noisy cloze", (mean +9.8 F1 across equivalent BERT-Base models). Unsupervised NMT question translation is also helpful, leading to a mean improvement of 1.8 F1 on BERT-Base for otherwise equivalent ``noisy cloze" models. The improvement over noisy clozes is surprisingly modest, and is discussed in more detail in Section \ref{discussion}.

\paragraph{Effect of QA model} BERT-Base is more effective than BiDAF+SA (an architecture specifically designed for QA). BERT-Large (not shown in Table \ref{ablations_table}) gives a further boost, improving our best configuration by 6.9 F1. 

\paragraph{Effect of Rule-based Generation} QA models trained on QA datasets generated by the Rule-based (RB) system of \citet{heilman_good_2010} do not perform favourably compared to our NMT approach. To test whether this is due to different answer types used, we a) remove questions of \emph{their} system that are not consistent with our (NE) answers, and b) remove questions of \emph{our} system that are not consistent with their answers.  
Table~\ref{rule_based_ablation_table} shows that while answer types matter in that using our restrictions help their system, and using their restrictions hurts ours, they cannot fully explain the difference. The RB system therefore appears to be unable to generate the \emph{variety} of questions and answers required for the task, \emph{and} does not generate questions from a sufficient \emph{variety} of contexts. Also, whilst on average, question lengths are shorter for the RB model than the NMT model, the distribution of longest common sequences are similar, as shown in Figure~\ref{question_lengths}, perhaps suggesting that the RB system copies a larger proportion of its input.

\begin{table}
\small
\centering
\begin{tabular}{ p{5.23cm} c c }
 \centering \textbf{Question Generation} & EM & F1 \\
\toprule
  \centering Rule Based  & 16.0 & 37.9\\
  \centering Rule Based (NE filtered) & 28.2 & 41.5\\
  \centering Ours & 38.6 & 47.8\\
 \centering  Ours (filtered for \context{},\answer{} pairs in Rule Based) & 38.5  & 44.7 \\
\end{tabular}
\caption{Ablations on SQuAD development set probing the performance of the rule based system. 
}
\label{rule_based_ablation_table}
\end{table}

\subsection{Error Analysis}
We find that the QA model predicts answer spans that are not always detected as named entity mentions (NEs) by the NER tagger, despite being trained with solely NE answer spans. In fact, when we split SQuAD into questions where the correct answer is an automatically-tagged NE, our model's performance improves to 64.5 F1, but it still achieves 47.9 F1 on questions which do not have automatically-tagged NE answers (not shown in our tables). We attribute this to the effect of BERT's linguistic pretraining allowing it to generalise the semantic role played by NEs in a sentence rather than simply learning to mimic the NER system. An equivalent BiDAF+SA model scores 58.9 F1 when the answer is an NE but drops severely to 23.0 F1 when the answer is not an NE.

\begin{figure}
\centering
  \includegraphics[width=0.475\textwidth]{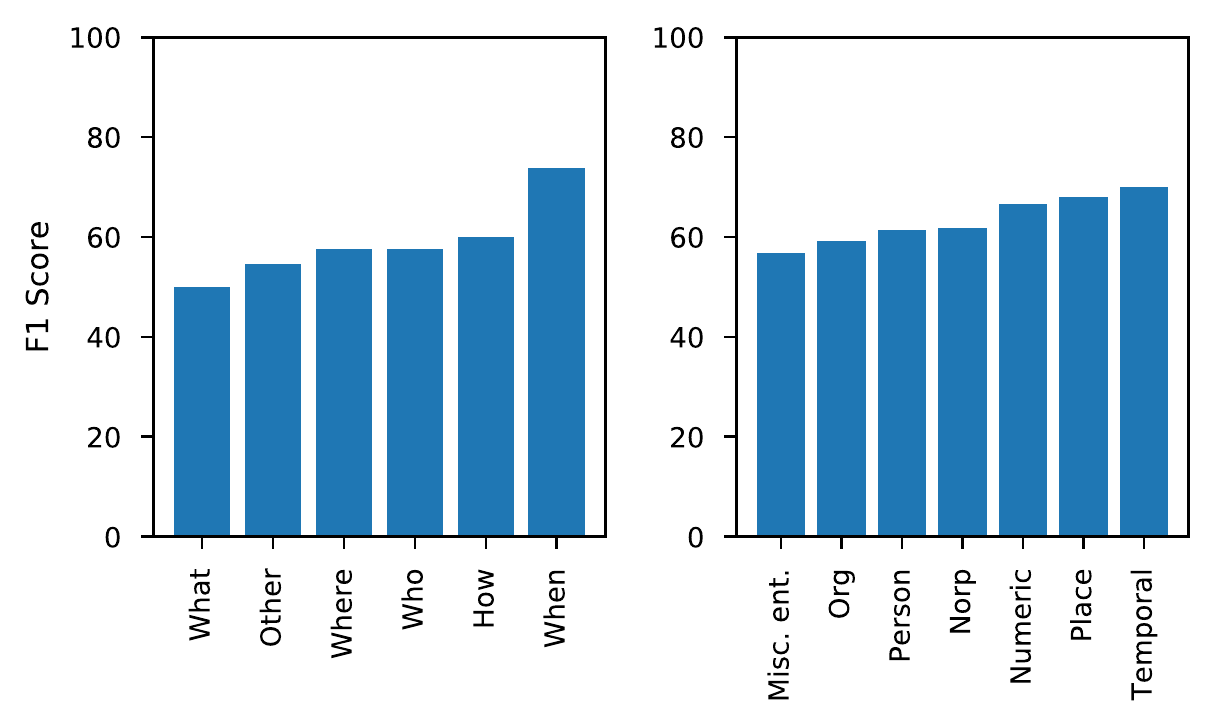}
  \caption{Breakdown of performance for our best QA model on SQuAD for different question types (left) and different NE answer categories (right)}
  \label{breakdown}
\end{figure}

Figure~\ref{breakdown} shows the performance of our system for different kinds of question and answer type. The model performs best with ``\texttt{when}" questions which tend to have fewer potential answers, but struggles with ``\texttt{what}" questions, which have a broader range of answer semantic types, and hence more plausible answers per context. The model performs well on ``\texttt{TEMPORAL}" answers, consistent with the good performance of ``\texttt{when}" questions.

\subsection{UNMT-generated Question Analysis}
\label{generated_question_analysis}
\begin{figure}
\centering
  \includegraphics[width=0.34\textwidth]{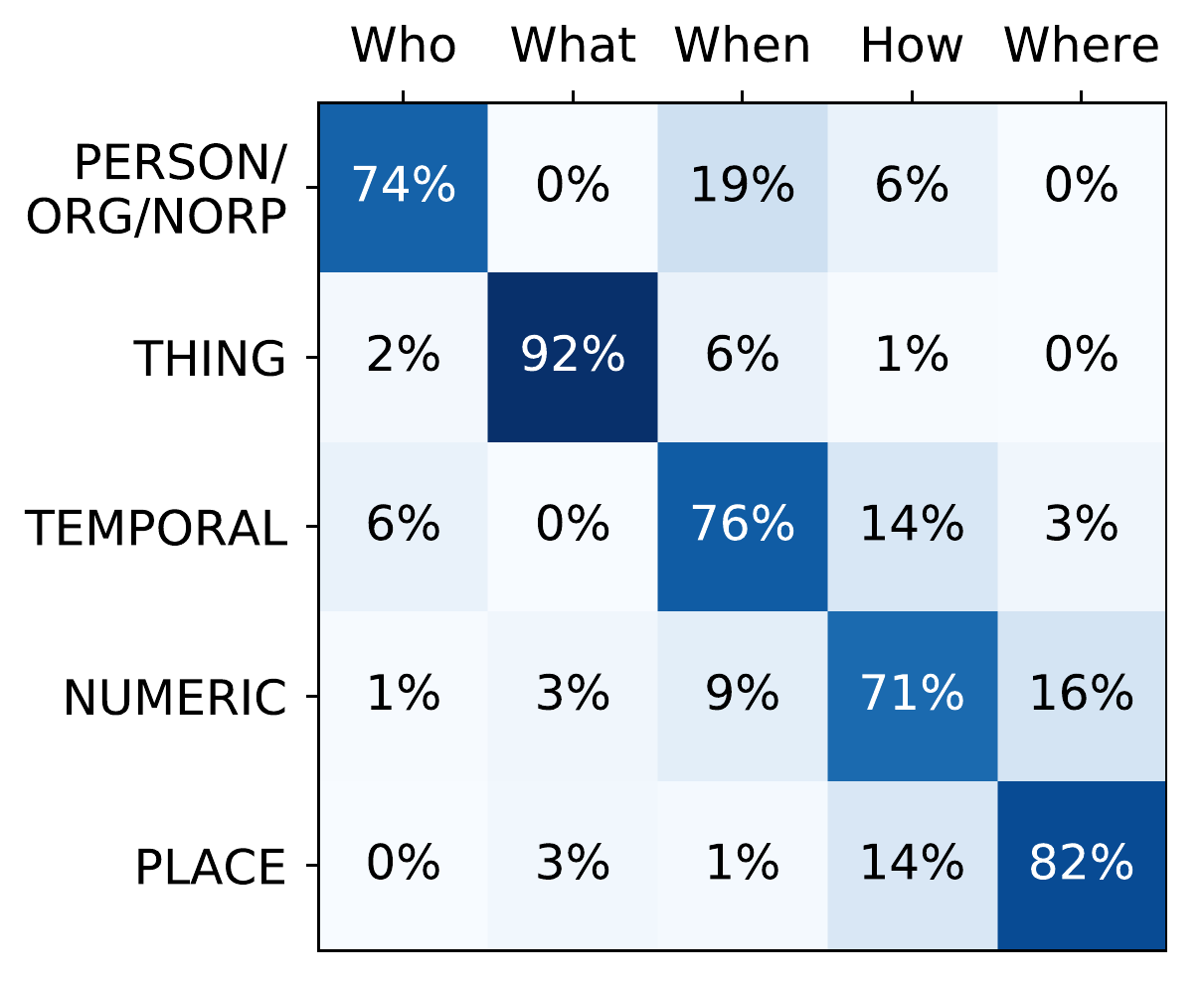}
  \caption{Wh* words generated by the UNMT model for cloze questions with different answer types. 
  }
  \label{wh_generation}
\end{figure}

Whilst our main aim is to optimise for downstream QA performance, it is also instructive to examine the output of the unsupervised NMT cloze translation system. Unsupervised NMT has been used in monolingual settings \cite{subramanian_multiple-attribute_2018}, but cloze-to-question generation presents new challenges -- The cloze and question are asymmetric in terms of word length, and successful translation must preserve the answer, not just superficially transfer style. Figure~\ref{wh_generation} shows that without the wh* heuristic, the model learns to generate questions with broadly appropriate wh* words for the answer type, but can struggle, particularly with Person/Org/Norp and Numeric answers.

Table~\ref{uqa_generations_table} shows representative examples from the NE unsupervised NMT model. The model generally copies large segments of the input.
Also shown in Figure \ref{question_lengths}, generated questions have, on average, a 9.1 token contiguous sub-sequence from the context, corresponding to 56.9\% of a generated question copied verbatim, compared to 4.7 tokens (46.1\%) for SQuAD questions. This is unsurprising, as the backtranslation training objective is to maximise the reconstruction of inputs, encouraging conservative translation.

The model exhibits some encouraging, non-trivial syntax manipulation and generation, particularly at the start of questions, such as example 7 in Table~\ref{uqa_generations_table}, where word order is significantly modified and ``sold" is replaced by ``buy". Occasionally, it hallucinates common patterns in the question corpus (example 6). The model can struggle with lists (example 4), and often prefers present tense and second person (example 5). Finally, semantic drift is an issue, with generated questions being relatively coherent but often having different answers to the inputted cloze questions (example 2).

We can estimate the quality and grammaticality of generated questions by using the well-formed question dataset of \citet{faruqui_identifying_2018}. This dataset consists of search engine queries annotated with whether the query is a well-formed question or not. We train a classifier on this task, and then measure how many questions are classified as ``well-formed" for our question generation methods. Full details are given in Appendix \ref{well_formed}. We find that 68\% of questions generated by UNMT model are classified as well-formed, compared to 75.6\% for the rule-based system and 92.3\% for SQuAD questions. We also note that using language model pretraining improves the quality of questions generated by UNMT model, with 78.5\% classified as well-formed, surpassing the rule-based system (see Appendix \ref{xlm}).

\begin{table*}
\footnotesize
\centering
\begin{tabular}{p{0.01cm} p{6.7cm}  p{1.15cm}  p{6cm} }
\# & \textbf{Cloze Question} & \textbf{Answer} & \textbf{Generated Question} \\   
\toprule
1 & they joined with PERSON/NORP/ORG to defeat him & Rom & Who did they join with to defeat him?\\
2 & the NUMERIC on Orchard Street remained open until 2009 & second & How much longer did Orchard Street remain open until 2009?\\
3 & making it the third largest football ground in PLACE & Portugal & Where is it making the third football ground? \\
4 & he speaks THING, English, and German & Spanish & What are we , English , and German? \\
5 & Arriving in the colony early in TEMPORAL & 1883 & When are you in the colony early? \\
6 & The average household size was NUMERIC & 2.30 & How much does a Environmental Engineering Technician II in Suffolk , CA make?\\
7 & WALA would be sold to the Des Moines-based PERSON/NORP/ORG for \$86 million & Meredith Corp & Who would buy the WALA Des Moines-based for \$86 million?
 \\

\end{tabular}
\caption{Examples of cloze translations for the UNMT model using the wh* heuristic and subclause cloze extraction. More examples can be found in appendix \ref{uqa_more_examples}}
\label{uqa_generations_table}
\end{table*}

\subsection{Few-Shot Question Answering}

Finally, we consider a few-shot learning task with very limited numbers of labelled training examples. We follow the methodology of \citet{dhingra_simple_2018} and \citet{yang_semi-supervised_2017}, training on a small number of training examples and using a development set for early stopping. We use the splits made available by \citet{dhingra_simple_2018}, but switch the development and test splits, so that the test split has n-way annotated answers. 
We first pretrain a BERT-large QA model using our best configuration from Section \ref{experimental}, then fine-tune  with a small amount of SQuAD training data. We compare this to our re-implementation of \citet{dhingra_simple_2018}, and training the QA model directly on the available data without unsupervised QA pretraining. 

\begin{figure}
\centering
  \includegraphics[width=0.48\textwidth]{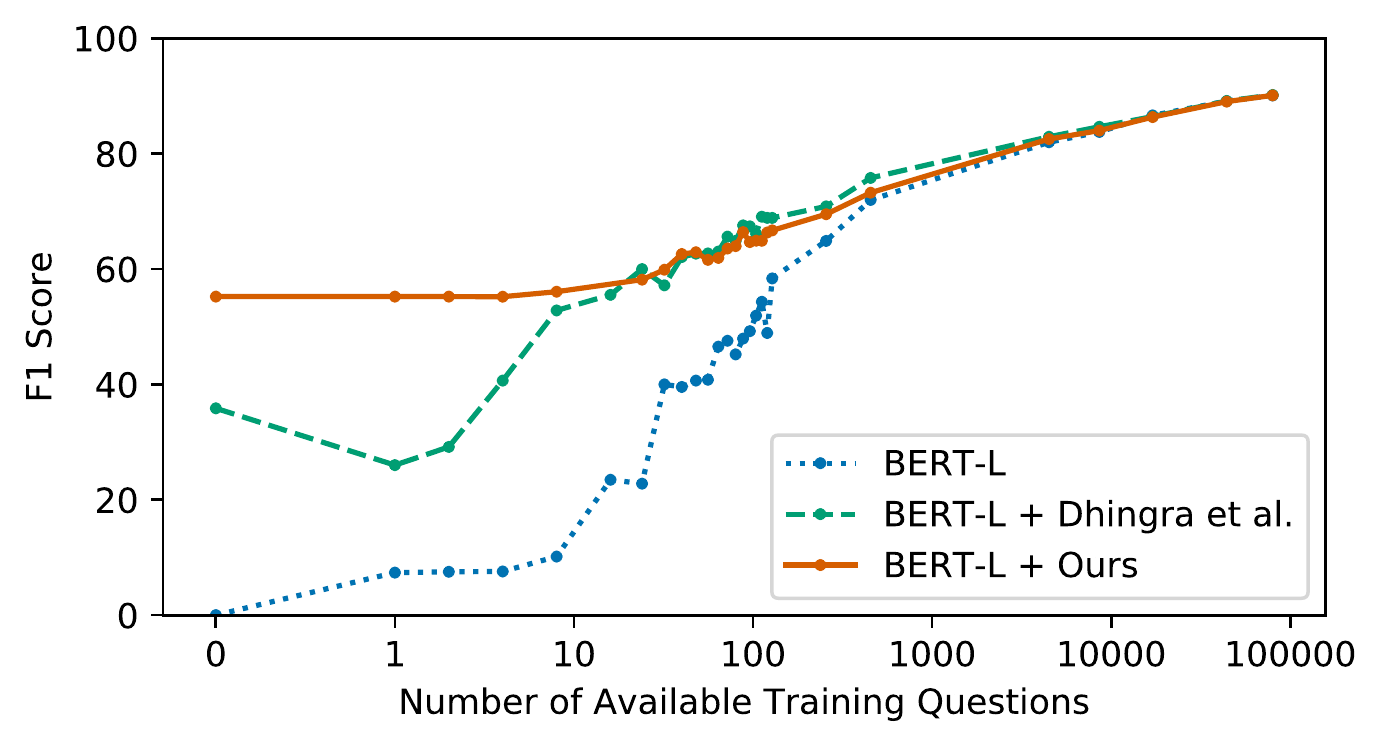}
  \caption{F1 score on the SQuAD development set for progressively larger training dataset sizes}
  \label{semi_fig}
\end{figure}

Figure~\ref{semi_fig} shows performance for progressively larger amounts of training data. As with \citet{dhingra_simple_2018}, our numbers are attained using a development set for early stopping that can be larger than the training set. Hence this is not a true reflection of performance in low data regimes, but does allow for comparative analysis between models. We find our approach performs best in very data poor regimes, and similarly to \citet{dhingra_simple_2018} with modest amounts of data. We also note BERT-Large itself is remarkably efficient, reaching $\sim$60\% F1 with only 1\% of the available data.

\begin{table*}
\centering
\begin{tabular}{ c | c | c | c || c  c | c  c }
\end{tabular}
\end{table*}

\section{Related Work}

\paragraph{Unsupervised Learning in NLP} Most representation learning approaches use latent variables \cite{Hofmann:1999:PLS:312624.312649,Blei:2003:LDA:944919.944937}, or language model-inspired criteria~\cite{Collobert,miko,glove,Radford, devlin_bert:_2018}. Most relevant to us is unsupervised NMT~\cite{muse,lample_unsupervised_2017,lample_phrase-based_2018,ArtetxeLA18} and style transfer~\cite{subramanian_multiple-attribute_2018}. We build upon this work, but instead of using models directly, we use them for training data generation. \citet{gpt2} report that very powerful language models can be used to answer questions from a conversational QA task, CoQA \cite{reddy_coqa:_2018} in an unsupervised manner. Their method differs significantly to ours, and may require ``seeding" from QA dialogs to encourage the language model to generate answers. \citet{yadav_alignment_2019} propose an unsupervised alignment method for multiple choice question answering.

\paragraph{Semi-supervised QA}
\citet{yang_semi-supervised_2017} train a QA model and also generate new questions for greater data efficiency, but require labelled data. \citet{dhingra_simple_2018} simplify the approach and remove the supervised requirement for question generation, but do not target unsupervised QA or attempt to generate natural questions. They also make stronger assumptions about the text used for question generation and require Wikipedia summary paragraphs. 
\citet{wang_multi-perspective_2018} consider semi-supervised cloze QA, \citet{chen_semi-supervised_2018} use semi-supervision to improve semantic parsing on WebQuestions \cite{berant_semantic_2013}, and \citet{lei_semi-supervised_2016} leverage semi-supervision for question similarity modelling. \citet{golub_two-stage_2017} propose a method to generate domain specific training QA instances for transfer learning between SQuAD and NewsQA \cite{yadav_alignment_2019}.
Finally, injecting external knowledge into QA systems could be viewed as semi-supervision, and \citet{weissenborn_dynamic_2017} and \citet{mihaylov_knowledgeable_2018} use Conceptnet \cite{speer_conceptnet_2016} for QA tasks.

\paragraph{Question Generation} has been tackled with pipelines of templates and syntax rules~\cite{rus_first_2010}. \citet{heilman_good_2010} augment this with a model to rank generated questions, and \citet{yao_semantics-based_2012} and \citet{olney_question_2012} investigate symbolic approaches. Recently there has been interest in question generation using supervised neural models, many trained to generate questions from $\context,\answer{}$ pairs in SQuAD \cite{du_learning_2017,yuan_machine_2017,zhao_paragraph-level_2018,du_harvesting_2018,hosking_evaluating_2019}

\section{Discussion}
\label{discussion}
It is worth noting that to attain our best performance, we require the use of both an NER system, indirectly using labelled data from OntoNotes 5, and a constituency parser for extracting sub-clauses, trained on the Penn  Treebank~\cite{marcus_penn_1994}.\footnote{Ontonotes 5: \url{https://catalog.ldc.upenn.edu/LDC2013T19}} 
Moreover, a language-specific wh* heuristic was used for training the best performing NMT models. 
This limits the applicability and flexibility of our best-performing approach to domains and languages that already enjoy extensive linguistic resources (named entity recognition and treebank datasets), as well as requiring some human engineering to define new heuristics. 

Nevertheless, our approach is unsupervised from the perspective of requiring no labelled (question, answer) or (question, context) pairs, which are usually the most challenging aspects of annotating large-scale QA training datasets. 

We note the ``noisy cloze" system, consisting of very simple rules and noise, performs nearly as well as our more complex best-performing system, despite the lack of grammaticality and syntax associated with questions. The questions generated by the noisy cloze system also perform poorly on the ``well-formedness" analysis mentioned in Section \ref{generated_question_analysis}, with only 2.7\% classified as well-formed. This intriguing result suggests natural questions are perhaps less important for SQuAD and strong question-context word matching is enough to do well, reflecting work from \citet{jia_adversarial_2017} who demonstrate that even supervised models rely on word-matching. 

Additionally, questions generated by our approach require no multi-hop or multi-sentence reasoning, but can still be used to achieve non-trivial SQuAD performance. Indeed, \citet{min_efficient_2018} note 90\% of SQuAD questions only require a single sentence of context, and \citet{sugawara_what_2018} find 76\% of SQuAD has the answer in the sentence with highest token overlap to the question. 

\section{Conclusion}

In this work, we explore whether it is possible to to learn extractive QA behaviour without the use of labelled QA data. We find that it is indeed possible, surpassing simple supervised systems, and strongly outperforming other approaches that do not use labelled data, achieving 56.4\% F1 on the popular SQuAD dataset, and 64.5\% F1 on the subset where the answer is a named entity mention. However, we note that whilst our results are encouraging on this relatively simple QA task, further work is required to handle more challenging QA elements and to reduce our reliance on linguistic resources and heuristics.

\section*{Acknowledgments}

The authors would like to thank Tom Hosking, Max Bartolo, Johannes Welbl, Tim Rockt{\"a}schel, Fabio Petroni, Guillaume Lample and the anonymous reviewers for their insightful comments and feedback.
\bibliography{acl2019}
\bibliographystyle{acl_natbib}

\clearpage

\twocolumn[
\centering
{\Large \emph{Supplementary Materials for ACL 2019 Paper:}}\\
\smallskip
{\Large Unsupervised Question Answering by Cloze Translation}\\
\bigskip
]
\appendix

\section{Appendices}
\label{sec:appendix}
\subsection{Cloze Question Featurization and Translation}
\label{cloze_feat}

Cloze questions are featurized as follows. Assume we have a cloze question extracted from a paragraph ``\emph{the Paris Sevens became the last stop on the calendar in \underline{\hspace{2em}}.}", and the answer ``\emph{2018}". We first tokenize the cloze question, and discard it if it is longer than 40 tokens. We then replace the ``blank" with a special mask token. If the answer was extracted using the noun phrase chunker, there is no specific answer entity typing so we just use a single mask token \texttt{"MASK"}. However, when we use the named entity answer generator, answers have a named entity label, which we can use to give the cloze translator a high level idea of the answer semantics. In the example above, the answer ``\emph{2018}" has the named entity type \texttt{"DATE"}. We group fine grained entity types into higher level categories, each with its own masking token as shown in Table~\ref{entity_map}, and so the mask token for this example is \texttt {"TEMPORAL"}.

\begin{table*}[h]
\centering
\small
\begin{tabular}{ p{3.9cm}  p{7.8cm}  p{3.0cm} }
\textbf{High Level Answer Category} & \textbf{Named Entity labels} & \textbf{Most appropriate wh*} \\
\toprule
PERSON/NORP/ORG & PERSON, NORP, ORG & Who\\
PLACE & GPE, LOC, FAC & Where\\
THING & PRODUCT, EVENT, WORKOFART, LAW, LANGUAGE & What \\
TEMPORAL & TIME, DATE & When\\
NUMERIC & PERCENT, MONEY, QUANTITY, ORDINAL, CARDINAL & How much/How many
\end{tabular}
\caption{High level answer categories for the different named entity labels}
\label{entity_map}
\end{table*}

\subsection{Unsupervised NMT Training Setup Details}
\label{uqa_training_details}

Here we describe experimental details for unsupervised NMT setup. We use the English tokenizer from Moses~\cite{koehn_moses:_2007}, and use FastBPE (\url{https://github.com/glample/fastBPE}) to split into subword units, with a vocabulary size of 60000. The architecture uses a 4-layer transformer encoder and 4-layer transformer decoder, where one layer is language specific for both the encoder and decoder, the rest are shared. We use the standard hyperparameter settings recommended by \citet{lample_phrase-based_2018}. The models are initialised with random weights, and the input word embedding matrix is initialised using FastText  vectors \cite{bojanowski_enriching_2016} trained on the concatenation of the $C$ and $Q$ corpora. Initially, the auto-encoding loss and back-translation loss have equal weight, with the auto-encoding loss coefficient reduced to $0.1$ by 100K steps and to $0$ by 300k steps.  We train using 5M cloze questions and natural questions, and cease training when the BLEU scores between back-translated and input questions stops improving, usually around 300K optimisation steps. When generating, we decode greedily, and note that decoding with a beam size of 5 did not significantly change downstream QA performance, or greatly change the fluency of generations.

\subsection{Wh* Heuristic}
\label{uqa_training_details_wh_heuristic}

We defined a heuristic to encourage appropriate wh* words for the inputted cloze question's answer type. This heuristic is used to provide a relevant wh* word for the ``noisy cloze" and ``identity" baselines, as well as to assist the NMT model to produce more precise questions. To this end, we map each high level answer category to the most appropriate wh* word, as shown on the right hand column of Table~\ref{entity_map} (In the case of \texttt{NUMERIC} types, we randomly choose between ``\texttt{How much}" and ``\texttt{How many}"). Before training, we prepend the high level answer category masking token to the start of questions that start with the corresponding wh* word, e.g. the question ``\emph{Where is Mount Vesuvius?}" would be transformed into ``\texttt{PLACE Where is Mount Vesuvius ?}". This allows the model to learn a much stronger association between the wh* word and  answer mask type.

\subsection{QA Model Setup Details}
\label{qa_model_training_details}

We train BiDAF + Self Attention using the default settings. We evaluate using a synthetic development set of data generated from 1000 context paragraphs every 500 training steps, and halt when the performance has not changed by 0.1\% for the last 5 evaluations.

We train BERT-Base and BERT-Large with a batch size of 16, and the default learning rate hyperparameters. For BERT-Base, we evaluate using a synthetic development set of data generated from 1000 context paragraphs every 500 training steps, and halt when the performance has not changed by 0.1\% for the last 5 evaluations. For BERT-Large, due to larger model size, training takes longer, so we manually halt training when the synthetic development set performance plateaus, rather than using the automatic early stopping.

\subsection{Question Well-Formedness}
\label{well_formed}

We can estimate how well-formed the questions generated by various configurations of our model are using the Well-formed query dataset of \citet{faruqui_identifying_2018}. This dataset consists of 25,100 search engine queries, annotated with whether the query is a well-formed question. We train a BERT-Base classifier on the binary classification task, achieving a test set accuracy of 80.9\% (compared to the previous state of the art of 70.7\%). We then use this classifier to measure what proportion of questions generated by our models are classified as ``well-formed". Table \ref{well_formed_table} shows the full results. Our best unsupervised question generation configuration achieves 68.0\%, demonstrating the model is capable of generating relatively well-formed questions, but there is room for improvement, as the rule-based generator achieves 75.6\%. MLM pretraining (see Appendix \ref{xlm}) greatly improves the well-formedness score. The classifier predicts that 92.3\% of SQuAD questions are well-formed, suggesting it is able to detect high quality questions. The classifier appears to be sensitive to fluency and grammar, with the ``identity" cloze translation models scoring much higher than their ``noisy cloze" counterparts. 

\begin{table}

\small
\centering
\begingroup
\setlength{\tabcolsep}{3pt}

\begin{tabular}{ c  c  c  c  c  }

\multirow{2}{30pt}{\centering \textbf{Cloze Answer}} & \multirow{2}{40pt}{\centering \textbf{Cloze Boundary}} & \multirow{2}{45pt}{\centering \textbf{Cloze Translation}} & \multirow{2}{36pt}{\centering \textbf{Wh* Heuristic}} & \multirow{2}{40pt}{\centering \textbf{\% Well-formed}} \\
 &&&&\\
  \toprule
  NE & Sub-clause & UNMT & \checkmark  & 68.0\\
  NE & Sub-clause & UNMT & $\times$    & 65.3 \\
  NE & Sentence   & UNMT & $\times$    & 61.3\\
  NP & Sentence   & UNMT & $\times$    & 61.9\\
  \midrule
  NE & Sub-clause & Noisy Cloze & \checkmark & 2.7\\
  NE & Sub-clause & Noisy Cloze & $\times$   & 2.4\\
  NE & Sentence   & Noisy Cloze & $\times$   & 0.7\\
  NP & Sentence   & Noisy Cloze & $\times$   & 0.8\\
  \midrule
  NE & Sub-clause & Identity & \checkmark & 30.8\\
  NE & Sub-clause & Identity & $\times$   & 20.0\\
  NE & Sentence   & Identity & $\times$   & 49.5\\
  NP & Sentence   & Identity & $\times$   & 48.0\\
  \midrule
    NE & Sub-clause & UNMT* & \checkmark  & \textbf{78.5}\\
\midrule
   \multicolumn{4}{c}{Rule-Based \cite{heilman_good_2010}} & 75.6\\
\midrule
   \multicolumn{4}{c}{SQuAD Questions \cite{rajpurkar_squad:_2016}} & \textbf{\emph{92.3}}\\

\bottomrule
\end{tabular}
\endgroup

\caption{Fraction of questions classified as "well-formed" by a classifier trained on the dataset of \citet{faruqui_identifying_2018} for different question generation models. * indicates MLM pretraining was applied before UNMT training}
\label{well_formed_table}
\end{table}

\subsection{Language Model Pretraining}
\label{xlm}

We experimented with Masked Language Model (MLM) pretraining of the translation models, $p_{s \rightarrow t}(\question | \cloze)$ and $p_{t \rightarrow s}(\cloze|\question)$. We use the XLM implementation (\url{https://github.com/facebookresearch/XLM}) and use default hyperparameters for both MLM pretraining and  and unsupervised NMT fine-tuning. The UNMT encoder is initialized with the MLM model's parameters, and the decoder is randomly initialized. We find translated questions to be qualitatively more fluent and abstractive than the those from the models used in the main paper. Table \ref{well_formed_table} supports this observation, demonstrating that questions produced by models with MLM pretraining are classified as well-formed 10.5\% more often than those without pretraining, surpassing the rule-based question generator of \citet{heilman_good_2010}. However, using MLM pretraining did not lead to significant differences for question answering performance (the main focus of this paper), so we leave a thorough investigation into language model pretraining for unsupervised question answering as future work.

\subsection{More Examples of Unsupervised NMT Cloze Translations}
\label{uqa_more_examples}
Table~\ref{uqa_generations_table} shows examples of cloze question translations from our model, but due to space constraints, only a few examples can be shown there. Table~\ref{uqa_more_generations_table} shows many more examples.

\begin{table*}
\footnotesize
\centering
\begin{tabular}{p{6.5cm} p{1.7cm} p{6.5cm} }
 \textbf{Cloze Question} & \textbf{Answer} & \textbf{Generated Question} \\
 \toprule
 to record their sixth album in TEMPORAL & 2005 & When will they record their sixth album ?\\
\midrule Redline management got word that both were negotiating with THING & Trek/Gary Fisher & What Redline management word got that both were negotiating ?\\
\midrule Reesler to suspect that Hitchin murdered PERSON/NORP/ORG & Wright & Who is Reesler to suspect that Hitchin murdered ?\\
\midrule joined PERSON/NORP/ORG in the 1990s to protest the Liberals' long-gun registry & the Reform Party & Who joined in the 1990s to protest the Liberals ' long-gun registry ?\\
\midrule to end the TEMPORAL NLCS, and the season, for the New York Mets & 2006 & When will the NLCS end , and the season , for the New York Mets ?\\
\midrule NUMERIC of the population concentrated in the province of Lugo & about 75\% & How many of you are concentrated in the province of Lugo ?\\
\midrule placed NUMERIC on uneven bars and sixth on balance beam & fourth & How many bars are placed on uneven bars and sixth on balance beam ?\\
\midrule to open a small branch in PLACE located in Colonia Escalon in San Salvador & La Casona & Where do I open a small branch in Colonia Escalon in San Salvador ?\\
\midrule they finished outside the top eight when considering only THING events & World Cup & What if they finished outside the top eight when considering only events ?\\
\midrule he obtained his Doctor of Law degree in 1929.Who's who in PLACE & America & Where can we obtain our Doctor of Law degree in 1929.Who ' s who ?\\
\midrule to establish the renowned Paradise Studios in PLACE in 1979 & Sydney & Where is the renowned Paradise Studios in 1979 ?\\
\midrule Ukraine came out ahead NUMERIC & four to three & How much did Ukraine come out ahead ?\\
\midrule their rule over these disputed lands was cemented after another Polish victory, in THING & the Polish-Soviet War & What was their rule over these disputed lands after another Polish victory , anyway ?\\
\midrule sinking PERSON/NORP/ORG 35 before being driven down by depth charge attacks & Patrol Boat & Who is sinking 35 before being driven down by depth charge attacks ?\\
\midrule to hold that PLACE was the sole or primary perpetrator of human rights abuses & North Korea & Where do you hold that was the sole or primary perpetrator of human rights abuses ?\\
\midrule to make it 2–1 to the Hungarians, though PLACE were quick to equalise & Italy & Where do you make it 2-1 to the Hungarians , though quick equalise ?\\
\midrule he was sold to Colin Murphy's Lincoln City for a fee of £NUMERIC & 15,000 & How much do we need Colin Murphy ' s Lincoln City for a fee ?\\
\midrule Bierut is the co-founder of the blog PERSON/NORP/ORG & Design Observer & Who is the Bierut co-founder of the blog ?\\
\midrule the Scotland matches at the 1982 THING being played in a "family atmosphere" & FIFA World Cup & What are the Scotland matches at the 1982 being played in a " family atmosphere " ?\\
\midrule Tom realizes that he has finally conquered both "THING" and his own stage fright & La Cinquette & What happens when Tom realizes that he has finally conquered both " and his own stage fright ?\\
\midrule it finished first in the PERSON/NORP/ORG ratings in April 1990 & Arbitron & Who finished it first in the ratings in April 1990 ?\\
\midrule his observer to destroy NUMERIC others & two & How many others can his observer destroy ?\\
\midrule Martin had recorded some solo songs (including "Never Back Again") in 1984 in PLACE & the United Kingdom & Where have Martin recorded some solo songs ( including " Never Back Again " ) in 1984 ?\\
\midrule the NUMERIC occurs under stadium lights & second & How many lights occurs under stadium ?\\
\midrule PERSON/NORP/ORG had made a century in the fourth match & Poulton & Who had made a century in the fourth match ?\\
\midrule was sponsored by the national liberal politician PERSON/NORP/ORG & Valentin Zarnik & Who was sponsored by the national liberal politician ?\\
\midrule Woodbridge also shares the PERSON/NORP/ORG with the neighboring towns of Bethany and Orange. & Amity Regional High School & Who else shares the Woodbridge with the neighboring towns of Bethany and Orange ?\\
\midrule A new Standard TEMPORAL benefit was introduced for university students & tertiary & When was a new Standard benefit for university students ?\\
\midrule mentions the Bab and THING & Bábís & What are the mentions of Bab ?\\
\end{tabular}
\caption{Further cloze translations from the UNMT model (with subclause boundaries and wh* heuristic applied)}
\label{uqa_more_generations_table}
\end{table*}

\end{document}